\pdfoutput=1

\documentclass[11pt]{article}

\usepackage[final]{acl}

\usepackage{times}
\usepackage{latexsym}

\usepackage[T1]{fontenc}

\usepackage[utf8]{inputenc}

\usepackage{microtype}

\usepackage{inconsolata}

\usepackage{graphicx}

\usepackage{multirow}
\usepackage{makecell}
\usepackage{colortbl}
\usepackage{setspace}
\usepackage{calligra}
\usepackage{algpseudocode}
\usepackage{arydshln}
\usepackage{tabu}
\usepackage{pifont}
\usepackage{tcolorbox}
\usepackage{subcaption}
\usepackage{booktabs}

\title{RealBench: A Chinese Multi-image Understanding Benchmark Close to Real-world Scenarios}

\author{
Fei Zhao$^{1,2}$, Chengqiang Lu$^2$, Yufan Shen$^3$, Qimeng Wang$^2$, Yicheng Qian$^2$, Haoxin Zhang$^2$, \\\textbf{Yan Gao}$^2$, \textbf{Yi Wu}$^2$, \textbf{Yao Hu}$^2$, \textbf{Zhen Wu}$^1${\thanks{\ \ Corresponding author.}}, \textbf{Shangyu Xing}$^1$, \textbf{Xinyu Dai}$^1$\\
$^1$National Key Laboratory for Novel Software Technology, Nanjing University\\
$^2$Xiaohongshu Inc. \quad $^3$Zhejiang University\\
{\tt \{zhaof,xingsy\}@smail.nju.edu.cn, syficy@zju.edu.cn,\{wuz,daixinyu\}@nju.edu.cn},\\
{\tt \{lusuo,sizhe1,jinqiu,haoli9,yadun,xiaohui,xiahou\}@xiaohongshu.com}
}

\begin{document}
\maketitle
\begin{abstract}
While various multimodal multi-image evaluation datasets have been emerged, but these datasets are primarily based on English, and there has yet to be a Chinese multi-image dataset. To fill this gap, we introduce RealBench, the first Chinese multimodal multi-image dataset, which contains 9393 samples and 69910 images. RealBench distinguishes itself by incorporating real user-generated content, ensuring high relevance to real-world applications. Additionally, the dataset covers a wide variety of scenes, image resolutions, and image structures, further increasing the difficulty of multi-image understanding. Ultimately, we conduct a comprehensive evaluation of RealBench using 21 multimodal LLMs of different sizes, including closed-source models that support multi-image inputs as well as open-source visual and video models. The experimental results indicate that even the most powerful closed-source models still face challenges when handling multi-image Chinese scenarios. Moreover, there remains a noticeable performance gap of around 71.8\% on average between open-source visual/video models and closed-source models. 
These results show that RealBench provides an important research foundation for further exploring multi-image understanding capabilities in the Chinese context. Our datasets will be available at \url{https://github.com/1429904852/RealBench}.


\end{abstract}

\section{Introduction}

In recent years, multimodal large language models (MLLMs) have seen remarkable advancements. Both closed-source models like GPT-4V~\cite{DBLP:journals/corr/abs-2303-08774}, Gemini~\cite{DBLP:journals/corr/abs-2312-11805}, Qwen-VL-Max~\cite{bai2023qwen}, and Claude-3.5-Sonnet~\cite{Anthropic_claude}, as well as open-source models such as LLaVA~\cite{DBLP:conf/nips/LiuLWL23a} and MiniGPT-4~\cite{DBLP:journals/corr/abs-2304-10592}, and BLIP~\cite{DBLP:conf/icml/0008LSH23}, have exhibited impressive visual-language understanding in single-image tasks such as  TextVQA~\cite{DBLP:conf/cvpr/SinghNSJCBPR19} and POPE~\cite{DBLP:conf/emnlp/LiDZWZW23}. However, real-world applications, particularly in social media platforms, predominantly involve multiple images presented simultaneously. For instance, many platforms feature posts containing 4-19 images to comprehensively convey information. This multi-image context poses unique challenges that extend beyond traditional single-image understanding, prompting a shift in research focus towards multi-image multimodal models~\cite{DBLP:conf/cvpr/LinYP0SH24,DBLP:journals/corr/abs-2312-14238} to better align with human cognitive processes in processing complex visual information.


To systematically evaluate multi-image understanding capabilities, researchers have developed various evaluation benchmarks, which can be categorized into three main approaches. Specifically, the majority of existing benchmarks ~\cite{DBLP:journals/corr/abs-2404-12390,DBLP:journals/corr/abs-2308-16463,DBLP:journals/corr/abs-2311-17092,DBLP:journals/corr/abs-2405-01483,DBLP:journals/corr/abs-2406-09411,DBLP:conf/iclr/0006PGG0ZCTZZ24,DBLP:journals/corr/abs-2404-18532,DBLP:journals/corr/abs-2406-12742,DBLP:journals/corr/abs-2408-02718,DBLP:conf/cvpr/YueNZ0LZSJRSWYY24,DBLP:conf/iclr/LuBX0LH0CG024} are constructed by first sampling images from existing vision datasets (e.g., IconQA~\cite{DBLP:conf/nips/LuQCXZZYLZ21}, NLVR2~\cite{DBLP:conf/acl/SuhrZZZBA19}), then generating corresponding multi-image tasks through either rule-based templates or LLM-assisted generation. A small portion of multi-image benchmarks are obtained from the English data on Wikipedia with the assistance of closed-source models like GPT-4o. The generated text is then manually checked to ensure its relevance to the associated images, as seen in datasets like MMDU~\cite{DBLP:journals/corr/abs-2406-11833}. Additionally, some multi-image benchmarks are entirely manually annotated, such as SlideVQA~\cite{DBLP:conf/aaai/TanakaNNHSS23}. Despite containing manually annotated QA pairs, SlideVQA is limited to slide-based content. Previous benchmarks, while cost-effective, may not fully capture the complexity and diversity of real-world multi-image scenarios.

\begin{figure*}
  \centering
  \begin{subfigure}{0.32\linewidth}
    \includegraphics[width=0.9\linewidth]{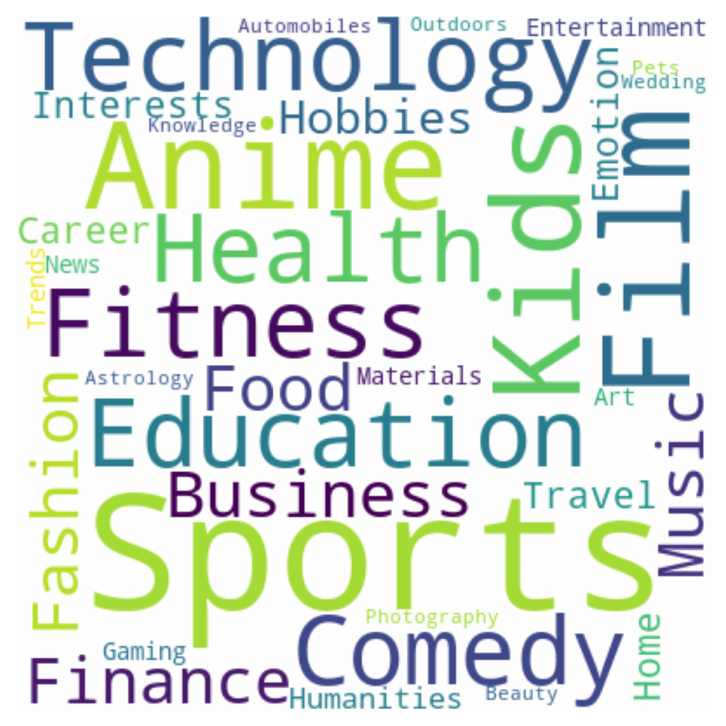}
    \caption{Word cloud of 36 categories.}
    \label{fig:data-scenaros}
  \end{subfigure}
  \begin{subfigure}{0.31\linewidth}
    \includegraphics[width=0.9\linewidth]{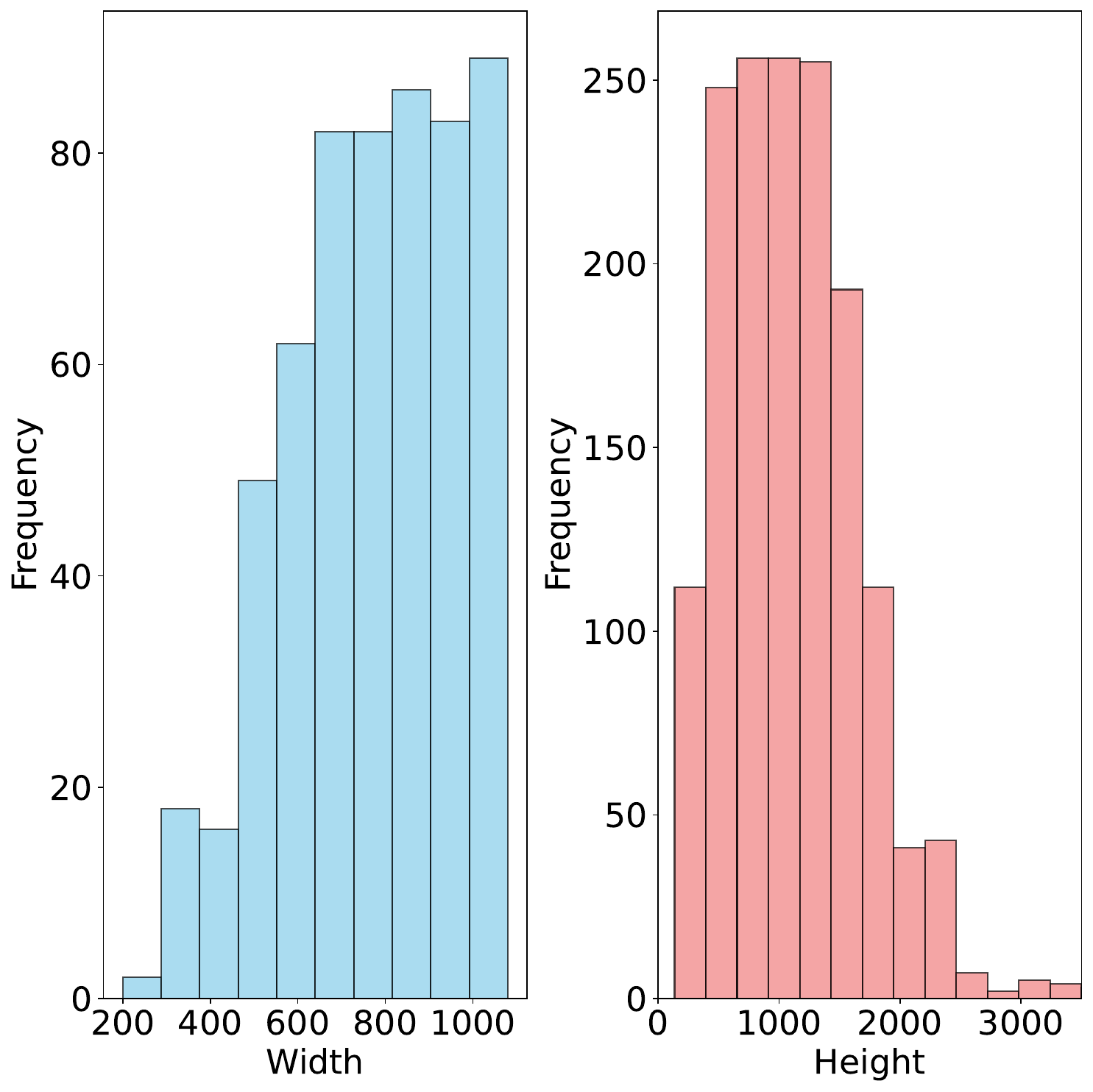}
    \caption{Distribution of width and height.}
    \label{fig:data-image}
  \end{subfigure}
  \begin{subfigure}{0.34\linewidth}
\includegraphics[width=0.9\linewidth]{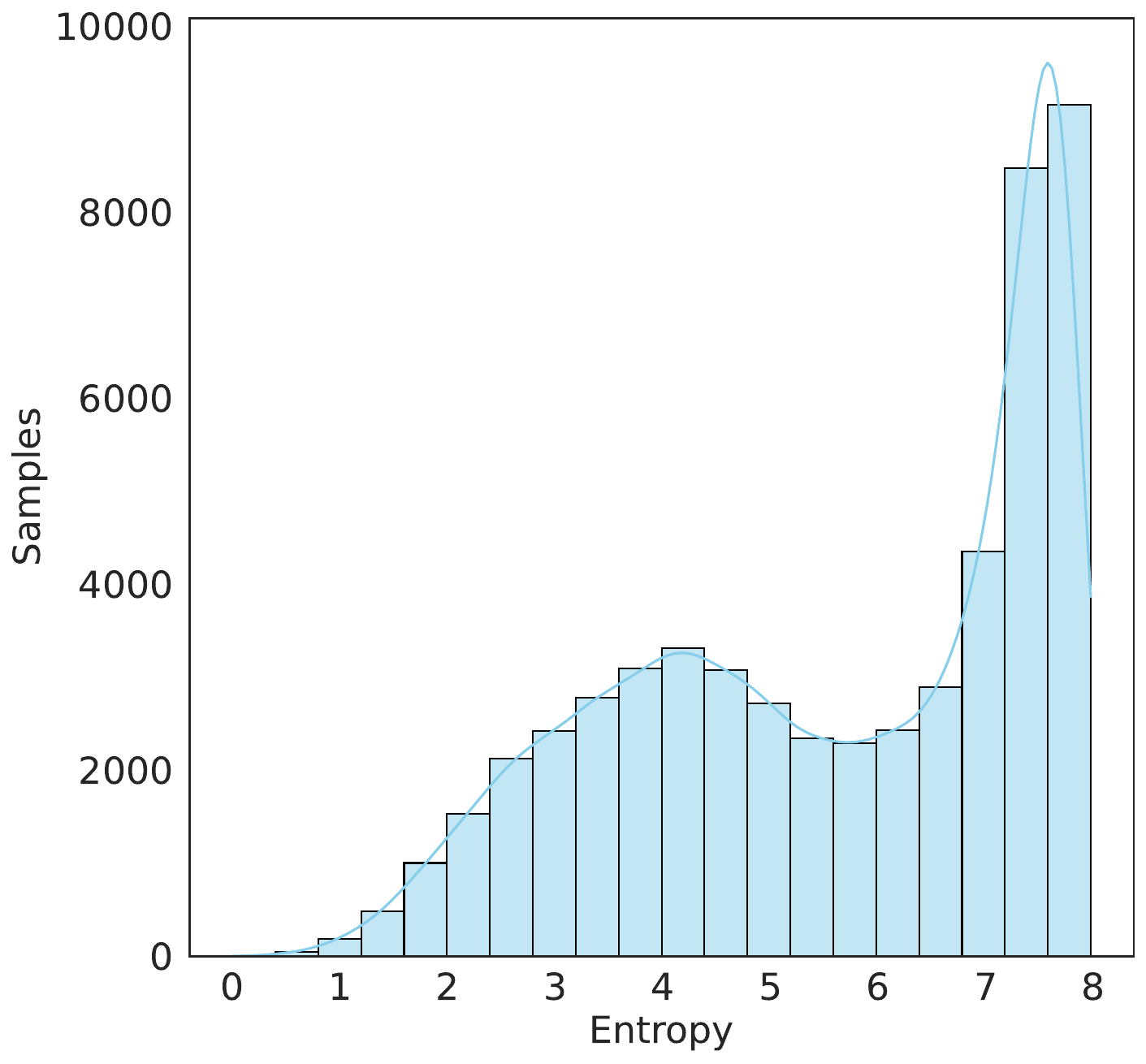}
    \caption{Distribution of image entropy.}
    \label{fig:data-entropy}
  \end{subfigure}
  \caption{The key characteristics of the RealBench. Figure (b) shows the distribution of image widths and heights. From the chart, we can see that the widths ranges from 200 to 1080, while the heights ranges from 137 to 3500, with a relatively scattered distribution, indicating the diversity of image resolutions. Figure (c) presents the distribution of image entropy, where image entropy is an important metric for measuring the amount of information in an image. Higher entropy values suggest that the image contains more details or complex structures, and thus has a greater information content. The entropy value ranges from 0 to 8, and as seen in Figure (c), the highest number of images corresponds to entropy values between 7 and 8, reflecting the diversity of image structures.}
  \label{fig:data_features}
\end{figure*}

Beyond the construction issue, existing datasets are predominantly English-centric, lacking Chinese multi-image understanding capabilities despite Chinese being widely used on social media. To address this, we introduce RealBench, the first Chinese multi-image benchmark built from public user-generated content. RealBench features three key characteristics (shown in Figure~\ref{fig:data_features}): (1) Diversity of Scenarios: The dataset encompasses 36 distinct categories drawn from everyday life scenarios, including fields such as travel experiences, food tutorials, home decoration, with a balanced distribution\footnote{The 36-categories distribution is shown in \textbf{Appendix}~\ref{Distribution_Categories}.}, fully reflecting the diversity of the real world; (2) Diversity of Resolutions: Due to users capturing images with different devices or under various shooting conditions in real-world scenarios, the images present natural variations in quality and resolution (ranging from 200 $\times$ 137 to 1080 $\times$ 3500 pixels); (3) Diversity of Image Structures: Given that the RealBench is directly derived from real-world environments, the image structures also exhibit diversity, including images with special fonts, images with mixed tables and text, and images containing nested sub-figures, all of which contribute to the increased complexity of multi-image understanding.

RealBench contains 9393 data samples and 69910 images, with an average of 7.4 images per sample. To systematically evaluate the model's multi-image understanding capability in real-world Chinese scenarios, we carefully design four complementary tasks that examine different aspects of visual-language processing: multi-image retrieval, multi-image ranking, multi-image extraction, and multi-image reasoning. These tasks are closely aligned with practical needs and are highly representative. Compared to previous multi-image tasks, each of which has unique challenges: (1) Multi-image Retrieval: Beyond conventional one-to-one matching, our one-to-many retrieval setting associates a single text with multiple images, aiming to evaluate whether the model can capture comprehensive semantic relationships across image groups; (2) Multi-image Ranking builds upon retrieval by introducing sequential understanding. With larger image sets (mean=6.2, max=18) than existing benchmarks (typically 3-4)~\cite{DBLP:journals/corr/abs-2311-17092,DBLP:conf/iclr/0006PGG0ZCTZZ24,DBLP:journals/corr/abs-2406-11833,DBLP:journals/corr/abs-2406-09411}, models must not only match content but also understand the logical or temporal order between images, simulating real-world scenarios like following step-by-step tutorials; (3) Multi-image Extraction further challenges models to identify and synthesize key information distributed across multiple images compared to a single image (one of multiple images) in previous work; (4) Multi-image Reasoning: We introduce complex reasoning tasks requiring logical inference across multiple images, extending beyond the simple single-image reasoning in existing benchmarks.

We conduct a comprehensive evaluation of RealBench using 21 multimodal LLMs with varying capabilities, including closed-source models that support multi-image inputs, as well as open-source visual and video models. Our evaluation reveals significant challenges in multi-image understanding, even for leading models. Specifically, we observe that: (1) Performance degradation in complex tasks: even top-performing models like GPT-4o and Claude-3.5-Sonnet achieve only 28.55\% and 39.32\% accuracy in one-to-many retrieval tasks, respectively; (2) Task difficulty progression: model performance consistently decreases as tasks become more complex, with reasoning tasks showing the largest gaps. For example, open-source visual model MiniCPM-V achieves an average Rough-L score of 0.51 in the multi-image extraction task, but this drops to around 0.24 in multi-image reasoning tasks, indicating that complex reasoning tasks place higher demands on the model's capabilities; (3) Cross-image integration bottleneck: models generally struggle when required to integrate information across multiple images. For example, in the multi-image extraction task, the leading model GPT-4o and Qwen-VL-Max has a Rough-L score of only 0.53, indicating a clear bottleneck in the model's ability to effectively integrate information from multiple images.

\begin{figure*}[t]
  \centering
   \includegraphics[width=0.98\linewidth]{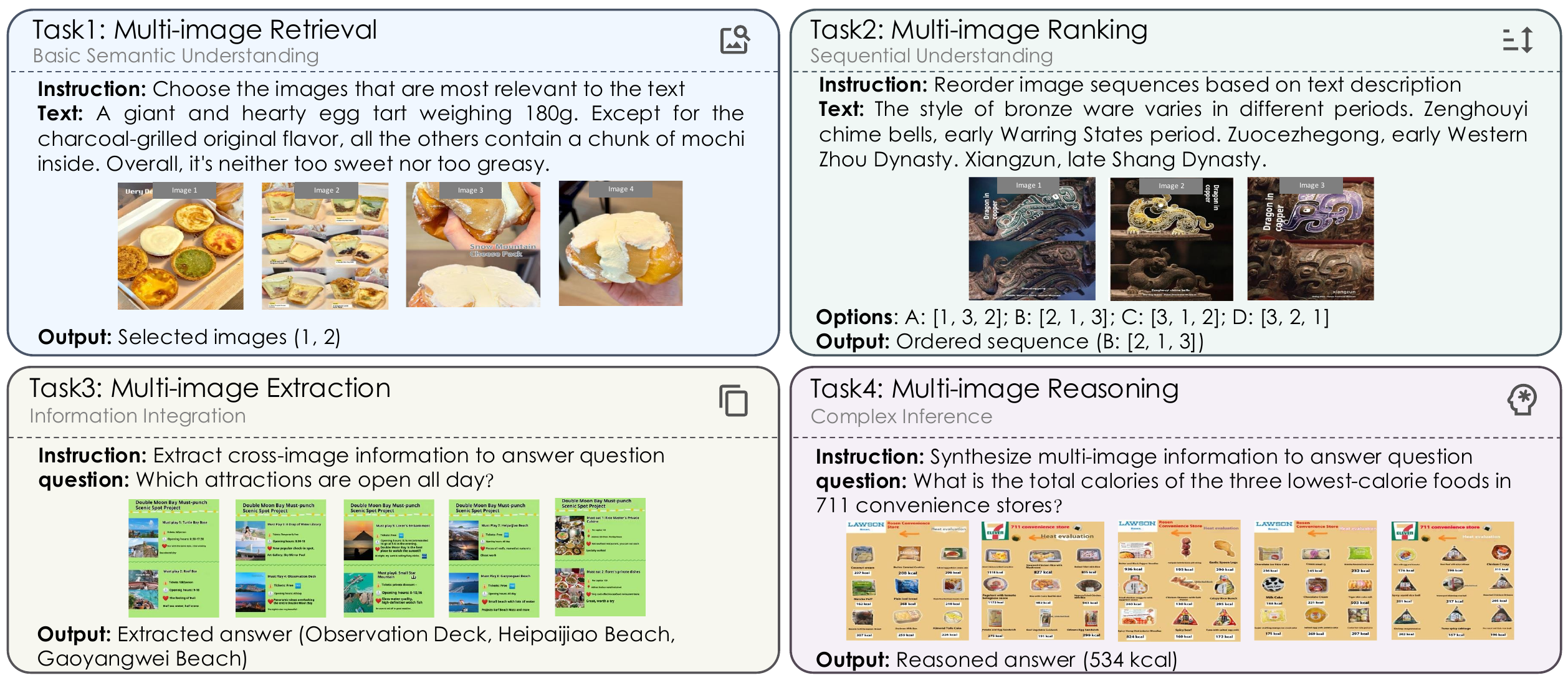}
   \caption{Detailed examples of the four tasks included in RealBench.}
   \label{fig:main}
\end{figure*}


\section{Related Work}

\subsection{Multimodal Large Language Models}

In recent years, multimodal large language models (MLLMs) have attracted increasing attention, with their core focus on achieving cross-modal understanding and generalization. Early MLLMs such as LLaVA~\cite{DBLP:conf/nips/LiuLWL23a}, MiniGPT-4~\cite{DBLP:journals/corr/abs-2304-10592}, BLIP~\cite{DBLP:conf/icml/0008LSH23}, AlignGPT~\cite{DBLP:journals/corr/abs-2405-14129}, and CogVLM~\cite{DBLP:journals/corr/abs-2311-03079} focused on single-image understanding, extracting visual cues and generating semantic outputs through integration with language models. These efforts laid the foundation for bridging the divide between visual and linguistic information.


While these models excel in single-image understanding, they show limitations in handling more complex multi-image scenarios. As multi-image understanding better aligns with human visual cognition, recent research has shifted towards multi-image MLLMs. To this end, models like MiniCPM~\cite{DBLP:journals/corr/abs-2408-01800}, Mantis~\cite{DBLP:journals/corr/abs-2405-01483}, InternVL~\cite{DBLP:journals/corr/abs-2312-14238}, and Qwen2-VL~\cite{wang2024qwen2} emerged to address the ability to process multiple image inputs. 


\subsection{Multi-Image Evaluation Benchmarks}
Researchers have proposed numerous single-image-based multimodal evaluation datasets. However, the development of multi-image evaluation datasets is still in its early stages. Recently, a series of multi-image benchmark tests have been launched successively. Most of them are created by combining or adapting existing open-source English datasets, with examples including BLINK~\cite{DBLP:journals/corr/abs-2404-12390}, Sparkles~\cite{DBLP:journals/corr/abs-2308-16463}, SEED-Bench~\cite{DBLP:journals/corr/abs-2311-17092}, Mantis~\cite{DBLP:journals/corr/abs-2405-01483}, MUIRBENCH~\cite{DBLP:journals/corr/abs-2406-09411}, DEMON~\cite{DBLP:conf/iclr/0006PGG0ZCTZZ24}, and MileBench~\cite{DBLP:journals/corr/abs-2404-18532}. This method effectively reduces development costs and speeding up the construction process. In addition, some multi-image datasets such as MMDU~\cite{DBLP:journals/corr/abs-2406-11833} are generated using data from Wikipedia, with the help of closed-source models like GPT-4V and GPT-4o. These datasets are manually reviewed to ensure that the generated content is relevant to the associated images. A very small number of multi-image datasets are constructed entirely through manual annotation, which typically requires a lot of manpower and time costs, such as SlideVQA~\cite{DBLP:conf/aaai/TanakaNNHSS23}.


Unlike existing English-centric datasets, RealBench introduces unique challenges in Chinese multi-image understanding by incorporating real-world user-generated content. The diverse image formats and complex layouts naturally occurring in Chinese social media platforms enable more robust evaluation of models' practical capabilities in handling Chinese visual-linguistic content.

\section{RealBench}

To evaluate the multi-image understanding capability of the model in real-world Chinese scenarios, we design four complementary tasks: retrieval, ranking, extraction, and reasoning. These tasks are commonly found in real-world scenarios, closely aligned with practical application needs, and are highly representative. Among them, the retrieval task ensures the efficient localization of relevant information, enhancing the speed of information acquisition; the ranking task optimizes the ordering of results to highlight key information; the extraction task accurately distills core content, improving the intuitiveness of the information; and the reasoning task performs in-depth analysis in complex situations to ensure the reliability of the results. Through these tasks, we can systematically evaluate the model's capabilities from multiple dimensions. Below is a detailed description of the task design and data collection process.


\subsection{Benchmark Construction}

Figure~\ref{fig:main} shows the details examples of the four tasks include in RealBench. Each of task has unique challenges. To ensure clarity for all reviewers, we provide English translations of the examples in Figure~\ref{fig:main}, while the original Chinese versions are available in \textbf{Appendix}~\ref{ChineseRealBench}.

\paragraph{Multi-image Retrieval.}

Multi-image retrieval aims to identify relevant images from a collection given a text query. While traditional approaches focus on one-to-one retrieval (matching single text to single image), real-world scenarios often require one-to-many retrieval, where multiple relevant images need to be identified for a given text. To this end, in addition to the traditional one-to-one retrieval method, we develop a one-to-many retrieval approach based on practical needs. To support the requirements of multi-image retrieval for these two scenarios, we construct different versions of the dataset, with one dataset serving as the easy version and the other as the hard version.


\begin{figure*}[t]
  \centering
   \includegraphics[width=0.99\linewidth]{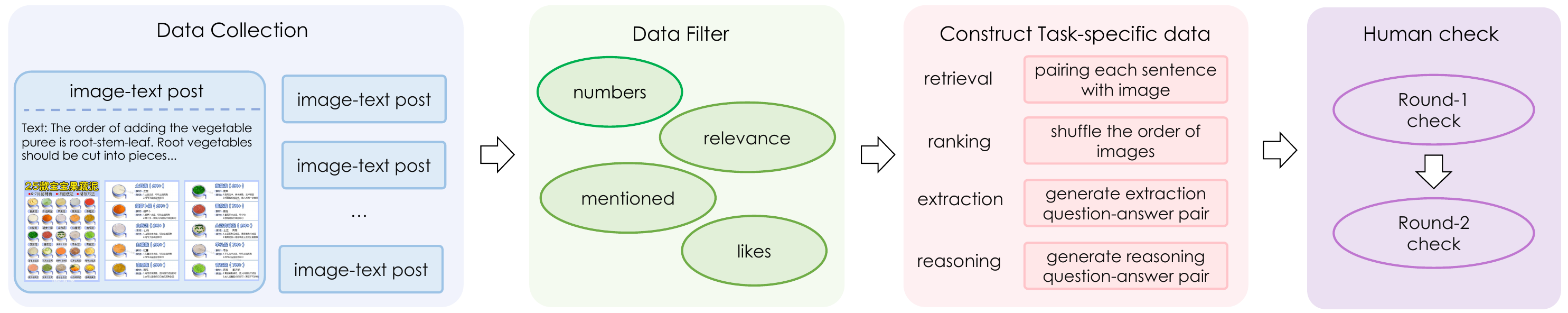}
   \caption{The flow of dataset construction and quality control.}
   \label{fig:data_construct_pipeline}
\end{figure*}

\paragraph{Multi-image Ranking.}

The goal of multi-image ranking is to reorder a shuffled list of images based on a given piece of text, resulting in a properly ordered list. The final answer includes four options, each representing a possible image arrangement, with the correct answer being one of these four options. Most prior multi-image datasets typically consist of an average of 3-4 images\cite{DBLP:journals/corr/abs-2311-17092,DBLP:conf/iclr/0006PGG0ZCTZZ24,DBLP:journals/corr/abs-2406-11833,DBLP:journals/corr/abs-2406-09411}, whereas our dataset has an average of 6 images, reaching up to 18 images at most. The increase in the number of images greatly raises the complexity of the ranking task. In light of this, we create two datasets with distinct levels of difficulty for the ranking task, based on the number of images involved. The first easy version dataset is designed for ranking up to 6 images (1-6 images), whereas the second hard version dataset focuses on ranking more than 6 images (7-18 images).



\paragraph{Multi-image Extraction.}

Multi-image extraction requires models to extract text-based answers from a collection of images in response to given queries. The answer must come from the text information in the image. In previous research, answers typically come from the text information within a single image (one of multiple images). However, there are instances where the answer requires consolidating text information from multiple images to be accurately addressed. Consequently, we introduce the requirement to extract multiple pieces of text information from several images as the correct answer, greatly increasing the difficulty of the task. In summary, apart from extracting answers from a single image, we also emphasize the need to extract answers from multiple images. To address both scenarios, we have specifically created two different versions of the dataset: a simple version and a hard version, to meet the requirements for multi-image extraction.



\paragraph{Multi-image Reasoning.}

Multi-image reasoning tasks require models to derive answers through logical inference based on textual information from multiple images. The task encompasses mainly four reasoning types\footnote{We provide an example for each type in \textbf{Appendix}~\ref{ChineseRealBench}.}: numerical calculation, counting, graphical reasoning, and attribute querying. While traditional approaches focus on single-image reasoning, we extend the task to cross-image reasoning, requiring information synthesis across multiple images. Similar to Multi-image Extraction, previous studies typically focus on reasoning based on a single image (one of multiple images), whereas we introduce the requirement to reason using the text from multiple images, which further enhances the challenges of the reasoning process. To this end, we construct two datasets with different difficulty levels — easy and hard. The first focuses on simple reasoning based on a single image (one of multiple images), while the second requires complex reasoning based on the text from multiple images. This innovation allows us to conduct a more comprehensive evaluation of the model's collaborative reasoning capabilities in multi-image scenarios.



\subsection{Data Collection and Quality Control}

In this subsection, we describe the pipeline of dataset construction and quality control, as shown in Figure~\ref{fig:data_construct_pipeline}.



\paragraph{Data Collection.}


In contrast to datasets that are adapted from existing ones or generated using Wikipedia data, our data is sourced directly from real user input. 
To be specific, we set some rough filtering criteria to select a batch of high-quality image-text posts from the Chinese social platform Xiaohongshu\footnote{\url{https://www.xiaohongshu.com}}: (1) the number of images in the image-text posts must be greater than one; (2) we select posts with a high image-text relevance score ($>$ 0.55); (3) we use regular expressions to select posts where the images are explicitly mentioned in the text, such as references like ``Figure 1'', ``Figure 2'', or ``Figure x''; (4) we select posts with a high number of likes, focusing primarily on posts with over 5000 likes to ensure the inclusion of popular and highly engaging content. After this rough filtering process, we obtain a high-quality candidate set. Each sample (i.e., each image-text post) in the candidate set contain not only a long text but also a corresponding set of images. The long text consists of multiple sentences and usually describes the content of a set of images in detail. This characteristic allows us to create a unique Chinese multi-image dataset, providing a realistic testing environment for multi-image tasks and greatly improves the practicality and value of the dataset.





\begin{figure}[t]
  \centering
   \includegraphics[width=0.65\linewidth]{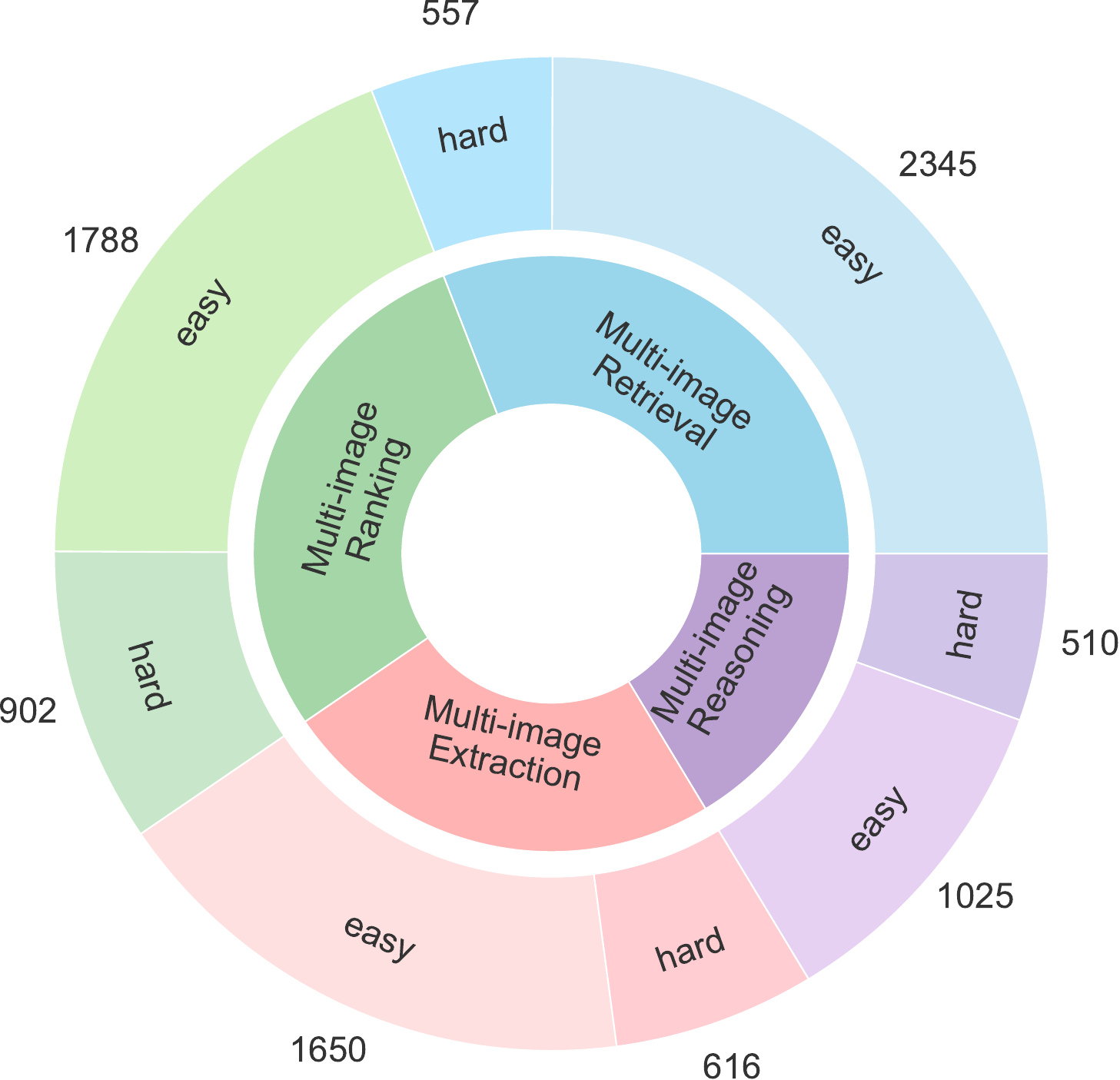}
   \caption{The dataset size of the four tasks.}
   \label{fig:dataset}
\end{figure}

\begin{table*}[t]
\setlength{\tabcolsep}{11pt}
\centering
\scalebox{0.63}{
\begin{tabu}{l|cc|cc|cc|cc|c}
\toprule
\multirow{2}{*}{\textbf{Method}} & \multicolumn{2}{c|}{\textbf{Multi-image Retrieval}} & \multicolumn{2}{c|}{\textbf{Multi-image Ranking}} & \multicolumn{2}{c|}{\textbf{Multi-image Extraction}} & \multicolumn{2}{c|}{\textbf{Multi-image Reasoning}} & \multirow{2}{*}{\textbf{Average}}\\
\multirow{-2}{*}{} & \textbf{easy} & \textbf{hard} & \textbf{easy} & \textbf{hard} & \textbf{easy} & \textbf{hard} & \textbf{easy} & \textbf{hard}\\
\midrule
\multicolumn{9}{c}{Closed-source MLLMs} \\
\midrule
GPT-4V & 41.07 & 14.00 & 25.00 & 23.61 & 33.55 & 28.48 & 13.58 & 12.79 & 24.01\\
GPT-4o & 65.33 & 28.55 & 23.38 & 25.28 & 67.85 & {\colorbox{green!20}{53.00}} & 33.32 & 28.74 & 40.68\\
Gemini-1.5-Flash & 48.96 & 11.67 & 25.28 & 26.39 & 30.42 & 37.10 & 7.38 & 10.14 & 24.67\\
Claude-3.5-Sonnet & 78.59 & {\colorbox{green!20}{39.32}} & 28.08 & {\colorbox{green!20}{28.60}} & 35.60 & 40.60 & 18.55 & 14.95 & 35.54\\
Qwen-VL-Max-0809 & {\colorbox{green!20}{82.94}} & 38.24 & {\colorbox{green!20}{31.26}} & 26.27 & {\colorbox{green!20}{72.69}} & {\colorbox{green!20}{53.00}} & {\colorbox{green!20}{39.21}} & {\colorbox{green!20}{31.43}} & {\colorbox{green!20}{46.88}}\\
\midrule
\multicolumn{9}{c}{Open-source MLLMs (Image models)} \\
\midrule
InternVL-1.5-26B & 5.54 & 0.36 & 14.10 & 12.40 & 2.91 & 4.57 & 4.26 & 7.02 & 6.40\\
InternVL-2.0-26B & 5.37 & 0.90 & 20.60 & 24.40 & 4.52 & 8.79 & 4.25 & 5.83 & 9.33\\
InternLM-XComposer-2.5 & 4.69 & 1.08 & 9.90 & 0.00 & 2.01 & 3.44 & 1.54 & 2.86 & 3.19\\
MiniCPM-V-2.6 & 59.02 & 12.03 & 25.62 & 25.28 & 60.50 & 41.34 & 31.29 & 17.49 & 34.07\\
Mantis-8B-Idefics2 & 31.34 & 0.36 & 23.99 & 20.18 & 2.80 & 3.06 & 0.65 & 0.58 & 10.37\\
Phi-3-vision-128k-instruct & 0.55 & 0.18 & 3.69 & 1.22 & 1.06 & 1.27 & 0.31 & 1.49 & 1.22\\
Qwen2-VL-7B & 14.54 & 0.90 & 25.06 & 23.61 & 9.30 & 14.20 & 6.46 & 6.24 & 12.54\\
Qwen2-VL-72B & 19.45 & 5.75 & 32.49 & 32.59 & 10.17 & 13.48 & 5.75 & 4.54 & 15.53\\
LLaVA-Next-Interleave & 5.37 & 0.90 & 25.06 & 21.29 & 9.11 & 12.98 & 4.91 & 6.90 & 10.82\\
LLaVa-Onevision & 1.49 & 0.18 & 26.62 & 25.83 & 6.84 & 9.41 & 3.56 & 3.90 & 9.73\\
Xgen-mm-interleave & 0.47 & 0.00 & 12.14 & 15.41 & 2.39 & 4.04 & 2.26 & 1.75 & 4.81\\
\midrule
\multicolumn{9}{c}{Open-source MLLMs (Video models)} \\
\midrule
VideoLLaMA2-7B & 0.09 & 0.18 & 24.61 & 22.17 & 4.40 & 13.69 & 6.11 & 10.98 & 10.28\\
Valley-13B & 0.00 & 0.00 & 0.00 & 0.00 & 2.42 & 3.47 & 0.49 & 0.19 & 0.82\\
CogVLM2-Video-LLama3 & 3.24 & 0.18 & 24.05 & 19.18 & 6.37 & 9.16 & 6.37 & 2.92 & 8.93\\
LongVila-8B & 2.17 & 0.00 & 0.34 & 0.11 & 0.00 & 0.00 & 0.20 & 0.00 & 0.35\\
LLaVa-Video & 33.43 & 11.85 & 25.95 & 26.16 & 12.27 & 16.42 & 3.30 & 3.00 & 16.55\\
\midrule
\textbf{Average} & 23.98 & 7.93 & 20.34 & 19.05 & 17.96 & 17.69 & 9.23 & 8.27 &\\
\bottomrule
\end{tabu}}
\caption{Main Results on RealBench. The best-performing results are highlighted in green for easy reference.}
\label{main_result}%
\end{table*}

\paragraph{Construction with GPT-4o.}

As mentioned above, each sample consists of a long text and a set of images. Next, we will introduce how to use closed-source models GPT-4o to assist in constructing the four multi-image tasks. For the multi-image retrieval task, we utilize a well-designed prompt to guide GPT-4o in pairing each sentence from the long text with the corresponding images in the image set. After establishing the correspondence between sentences and images, we further transformed it into a multi-modal multi-image retrieval task. To construct the multi-image ranking task, we first randomly shuffle the order of images within each sample. Next, we build the correct image ranking based on the long text description in each sample. In terms of answer construction, we use a multiple-choice format that not only includes the correct image ranking option but also features three incorrect ranking options to enhance the challenge of the task. In conducting multi-image extraction and multi-image reasoning, we need to fully utilize the textual information in the images. Therefore, we first filter out images with OCR text counts of fewer than 10 characters to ensure that the data we use contains sufficient content. Next, we design different prompts for these two tasks to better guide the model's output. Finally, we input the set of images from each sample into GPT-4o, using the prompts to construct question-answer pairs that meet the task requirements. The prompts\footnote{Each task has two datasets of different difficulty levels, so we design two distinct prompts for each task.} used for data construction can be found in \textbf{Appendix}~\ref{Prompt_Format}.



\paragraph{Quality Control with Human Annotators.}

To ensure the quality of the data, we implement strict quality control measures for the data generated by GPT-4o, taking two key steps: (1) each data sample must be reviewed by five different annotators\footnote{The annotators are university students, and the cost for each annotator is approximately \$55 per day.}, with all annotators conducting blind assessments to maintain the independence and objectivity of the process. Only when all five agree on its usability can it be considered as candidate data; (2) the approved candidate data undergoes a secondary quality check completed by three experts to ensure it meets the established standards and expectations. These measures not only enhance the overall credibility of the data but also lay a solid foundation for subsequent research and applications. Finally, RealBench contains a total of 9393 data samples and 69910 images. The dataset sizes for each of the four tasks are presented in Figure~\ref{fig:dataset}.

\section{Experiments}

\subsection{Experimental Setup}

\paragraph{Evaluation Models.}


We perform a comprehensive evaluation of 21 recent multimodal LLMs on RealBench, covering models specifically designed for multi-image inputs. To better analyze their performance and characteristics, we categorize these models into three distinct groups: five closed-source models (GPT-4V~\cite{DBLP:journals/corr/abs-2303-08774}, GPT-4o, Gemini-1.5-Flash~\cite{DBLP:journals/corr/abs-2312-11805}, Claude-3.5-Sonnet~\cite{Anthropic_claude}, Qwen-VL-Max-0809~\cite{bai2023qwen}), eleven open-source image models (InternVL-1.5-26B~\cite{DBLP:journals/corr/abs-2404-16821}, InternVL-2.0-26B, InternLM-XComposer-2.5~\cite{DBLP:journals/corr/abs-2407-03320}, MiniCPM-V-2.6~\cite{DBLP:journals/corr/abs-2408-01800}, Mantis-8B-Idefics2~\cite{DBLP:journals/corr/abs-2405-01483}, Phi-3-vision-128k-instruct~\cite{DBLP:journals/corr/abs-2404-14219}, Qwen2-VL-7B~\cite{wang2024qwen2}, Qwen2-VL-72B, LLaVA-Next-Interleave~\cite{DBLP:journals/corr/abs-2407-07895}, LLaVa-Onevision~\cite{DBLP:journals/corr/abs-2408-03326}, Xgen-mm-interleave~\cite{DBLP:journals/corr/abs-2408-08872}) and five open-source video models (VideoLLaMA2-7B~\cite{DBLP:journals/corr/abs-2406-07476}, Valley-13B~\cite{DBLP:journals/corr/abs-2306-07207}, CogVLM2-Video-LLama3~\cite{DBLP:journals/corr/abs-2408-16500}, LongVila-8B~\cite{DBLP:journals/corr/abs-2408-10188}, LLaVa-Video~\cite{zhang2024llava}).

\paragraph{Evaluation setup.} All evaluations are conducted exclusively in a zero-shot manner. The evaluation prompts corresponding to each task is detailed in \textbf{Appendix}~\ref{Prompt_Format}. For a fair comparison, the temperature is set to 0 and retry is set to 5. Multi-image extraction and reasoning are generative tasks, and following prior research~\cite{DBLP:journals/corr/abs-2405-01483,DBLP:conf/iclr/0006PGG0ZCTZZ24,DBLP:journals/corr/abs-2404-18532}, we utilize the widely recognized n-gram-based metric, ROUGE-L, to assess performance. We also incorporate human evaluation to ensure the reliability of results. In contrast, for multi-image retrieval and ranking tasks, accuracy is chosen as the primary evaluation metric to measure effectiveness. We conduct all experiments on NVIDIA A100 GPUs.



\subsection{Main Result}

The main experiment results are shown in Table~\ref{main_result}, from which we have following findings:

\paragraph{Closed-source models generally perform better than open-source image and video models.}

The experimental results reveal that even the most powerful closed-source models face considerable difficulties when evaluate on the four Chinese multi-image tasks. In many cases, these models struggle to reach acceptable performance levels, particularly in multi-image ranking and multi-image reasoning tasks. For example, the ROUGE-L of GPT-4o and Claude-3.5-Sonnet in multi-image reasoning is only 33.32\% and 18.55\%, respectively. It is worth noting that Qwen-VL-Max-0809 performs the best among the closed-source models, but even so, it still fails to achieve satisfactory results in certain tasks. These results underscore the substantial difficulty posed by our datasets, highlighting their challenge for current state-of-the-art models. Compared to closed-source models, the overall performance of open-source image and video models is notably weaker. As we can see, aside from MiniCPM-V-2.6, which is comparable to the closed-source models in terms of performance, all other open-source models fall behind. We speculate that the limited performance of current MLLMs on our benchmark can be attributed to the following factors: (1) Multi-image reasoning complexity: Our benchmark requires understanding not just individual images, but also their temporal or logical order, and performing cross-image integration and reasoning. Many MLLMs struggle to integrate and align information across multiple inputs.
(2) Real-world image complexity: Unlike synthetic datasets, our images are collected from real-world user-generated content. They contain diverse visual styles, layouts, embedded texts, and varying resolutions, which present significant challenges in robust visual understanding. (3) Language-specific limitations: The benchmark is entirely in Chinese, while many strong-performing MLLMs have been primarily trained and evaluated on English data. This language mismatch further impacts their performance, especially in tasks that require fine-grained alignment between Chinese text and images. These findings suggest that open-source models still have relatively limited capabilities in Chinese multi-image understanding tasks, leaving substantial room for improvement and future advancement.




\begin{figure}[t]
  \centering
   \includegraphics[width=0.95\linewidth]{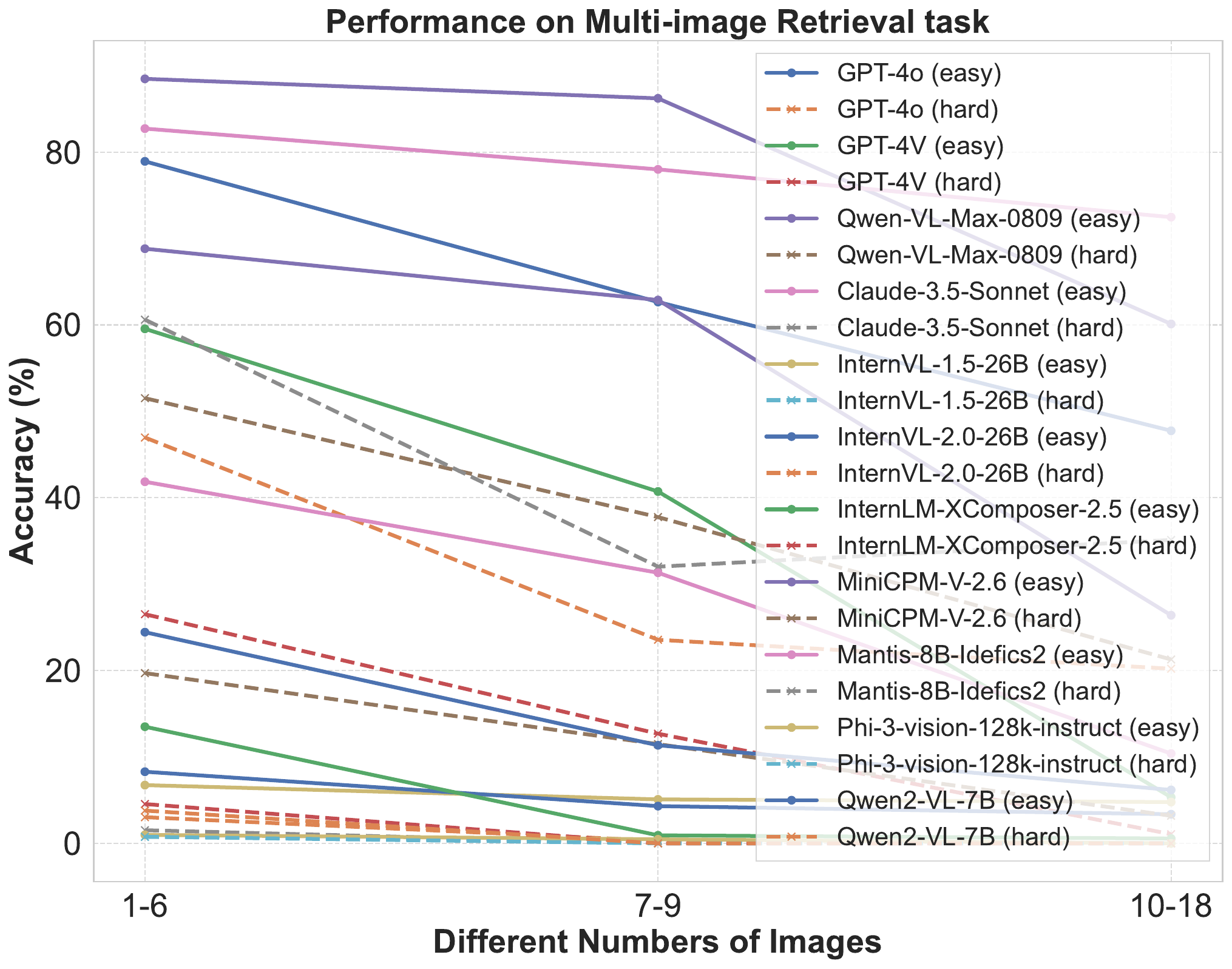}
   \caption{The performance of MLLMs on the multi-image retrieval task with different numbers of images.}
   \label{fig:image_number}
\end{figure}

\paragraph{The model typically performs better on the easy dataset than on the hard dataset.}


As mentioned before, each multi-image task has two versions: easy and hard. The two versions of the dataset differ in size, making direct comparisons somewhat unfair. Moreover, the averaging results may be skewed by model selection. Nevertheless, this metric still holds a certain reference value in revealing overall performance trends. Therefore, we consider this metric as a supplementary reference only. Both open-source and closed-source models generally perform better on the easy dataset than on the hard dataset. As shown in the last row of Table~\ref{main_result}, for each task, the average performance on the hard version is lower than that on the easy version. Additionally, in the multi-image retrieval task, 19 models perform better on the easy dataset than on the hard dataset; in the multi-image reasoning task, 13 models outperform on easy compared to hard. This difference is reasonable, as the hard dataset not only requires understanding multiple images but also demands deeper analysis and reasoning capabilities from the models.

\begin{table}[t]
\setlength{\tabcolsep}{13pt}
\centering
\scalebox{0.60}{
\begin{tabu}{l|cc|cc}
\toprule
\multirow{2}{*}{\textbf{Method}} & \multicolumn{2}{c|}{\textbf{Task2 (Regular)}} & \multicolumn{2}{c}{\textbf{Task2 (Shuffled)}}\\
\multirow{-2}{*}{} & \textbf{easy} & \textbf{hard} & \textbf{easy} & \textbf{hard}\\ 
\midrule
\multicolumn{5}{c}{Closed-source MLLMs} \\
\midrule
GPT-4V & 25.00 & 23.61 & 26.79 & 27.27\\
GPT-4o & 23.38 & 25.28 & 25.84 & 26.61\\
Claude-3.5-Sonnet & 28.08 & 28.60 & 26.73 & 26.16\\
Qwen-VL-Max-0809 & 31.26 & 26.27 & 31.66 & 23.95\\
\midrule
\multicolumn{5}{c}{Open-source MLLMs (Image models)} \\
\midrule
InternVL-1.5-26B & 14.10 & 12.40 & 12.75 & 9.31\\
InternVL-2.0-26B & 20.60 & 24.40 & 20.75 & 20.07\\
InternLM-XComposer-2.5 & 9.90 & 0.00 & 7.44 & 0.00\\
MiniCPM-V-2.6 & 25.62 & 25.28 & 20.81 & 20.84\\
Mantis-8B-Idefics2 & 23.99 & 20.18 & 22.54 & 17.52\\
Phi-3-vision-128k-instruct & 3.69 & 1.22 & 4.59 & 0.78\\
Qwen2-VL-7B & 25.06 & 23.61 & 26.01 & 26.94\\
\midrule
\multicolumn{5}{c}{Open-source MLLMs (video models)} \\
\midrule
VideoLLaMA2-7B & 24.61 & 22.17 & 24.66 & 27.83\\
Valley-13B & 0.00 & 0.00 & 0.00 & 0.00\\
CogVLM2-Video-LLama3 & 24.05 & 19.18 & 25.95 & 20.84\\
LongVila-8B & 0.34 & 0.11 & 0.34 & 0.00\\
\bottomrule
\end{tabu}}
\caption{The performance of MLLMs on the multi-image ranking during the data contamination evaluation.}
\label{data_risk}%
\end{table}

\subsection{Ablation Analysis}

\paragraph{The performance of MLLMs under different numbers of images.}

According to the statistics, each sample in the dataset contains up to 18 images. To study the performance variation of MLLMs with different numbers of images, we divide the dataset into three levels based on the number of images per sample: small, medium, and large. Specifically, these levels correspond to 1-6 images, 7-9 images, and 10-18 images, respectively. Without loss of generality, we chose the multi-image retrieval task for this exploration. Figure~\ref{fig:image_number} shows the performance trends of the models across these three categories with varying numbers of images. It can be observed that as the number of images increases, the performance of both open-source and closed-source models declines significantly. This is primarily due to the increasing complexity of tasks as the number of images grows, making it more challenging for the models to process and thus impacting their overall performance.

\paragraph{The risk of data contamination in RealBench.}

We conduct an in-depth investigation to assess whether there is any potential data contamination in the dataset. Specifically, we implement a method where the four options in the multi-image ranking task are randomly shuffled, ensuring that the newly arranged correct option was different from the original correct option. Afterward, we have all models re-predict the correct answer to evaluate their performance under the new arrangement. The results are shown in Table~\ref{data_risk}. The final results show that all models maintain stable performance, with only minimal fluctuations. This indicates that none of the models had been pre-trained on our dataset, and the likelihood of data contamination is extremely low. Thus, we have reason to believe that the models' performance in the experiment reflects their true ability to handle the tasks, rather than being skewed by data contamination.

\begin{figure}[t]
  \centering
   \includegraphics[width=0.98\linewidth]{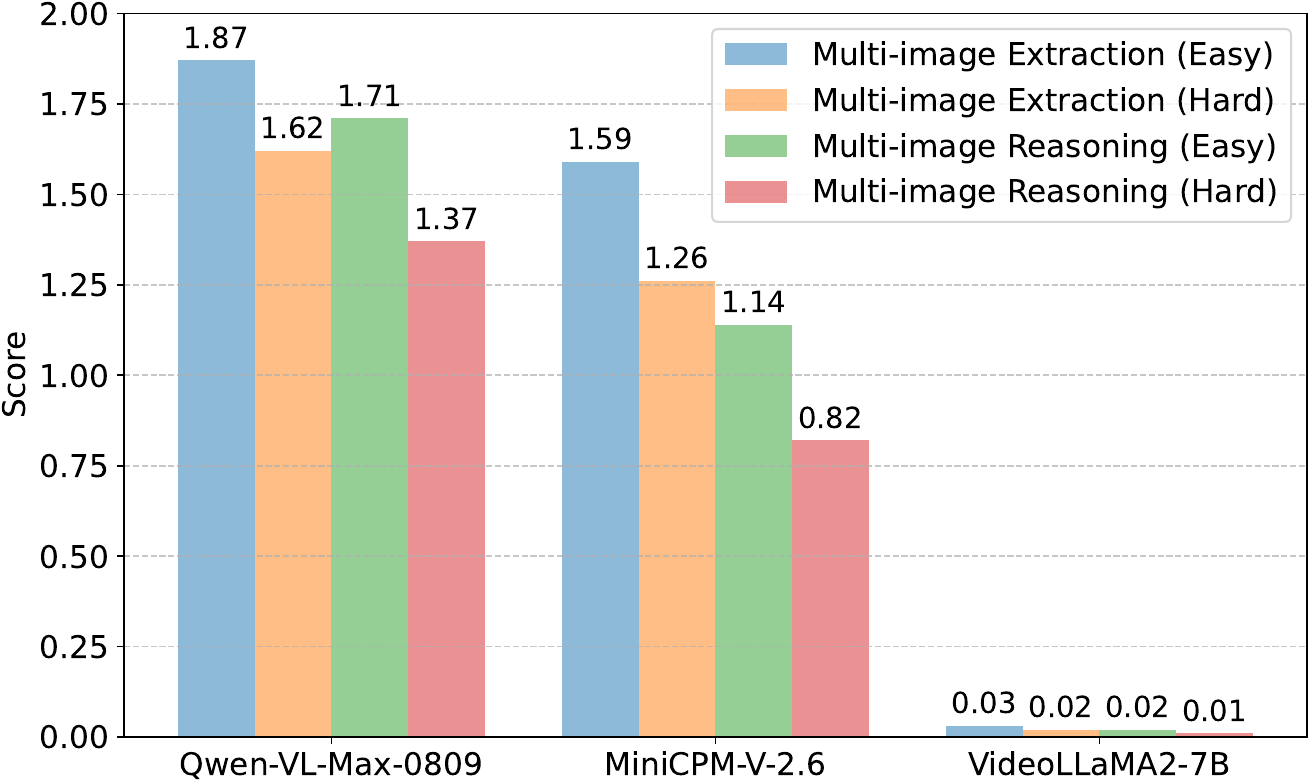}
   \caption{Human evaluation results of multi-image extraction and multi-image reasoning tasks.}
   \label{table:Human_Evaluation_result}%
\end{figure}

\begin{table*}[t]
\centering
\scalebox{0.70}{
\begin{tabu}{l|cc|cc|cc|cc|c}
\toprule
\multirow{2}{*}{\textbf{Method}} & \multicolumn{2}{c|}{\textbf{Multi-image Retrieval}} & \multicolumn{2}{c|}{\textbf{Multi-image Ranking}} & \multicolumn{2}{c|}{\textbf{Multi-image Extraction}} & \multicolumn{2}{c|}{\textbf{Multi-image Reasoning}}& \multirow{2}{*}{\textbf{Average}}\\
\multirow{-2}{*}{} & \textbf{easy} & \textbf{hard} & \textbf{easy} & \textbf{hard} & \textbf{easy} & \textbf{hard} & \textbf{easy} & \textbf{hard}\\ 
\midrule
\multicolumn{9}{c}{Closed-source MLLMs} \\
\midrule
GPT-4V & 42.13 & 12.21 & 25.78 & 24.50 & 27.96 & 25.14 & 12.44 & 12.43 & 22.82\\
GPT-4o & {\colorbox{green!20}{71.81}} & {\colorbox{green!20}{30.52}} & 23.66 & 22.95 & {\colorbox{green!20}{48.87}} & {\colorbox{green!20}{42.20}} & 21.70 & {\colorbox{green!20}{22.53}} & {\colorbox{green!20}{35.53}}\\
Claude-3.5-Sonnet & 63.75 & 28.9 & {\colorbox{green!20}{32.83}} & {\colorbox{green!20}{30.16}} & 17.41 & 25.72 & 11.77 & 12.42 & 14.14\\
Qwen-VL-Max-0809 & 7.76 & 0.18 & 23.71 & 25.17 & 19.03 & 19.08 & 11.65 & 6.50 & 27.87\\
\midrule
\multicolumn{9}{c}{Open-source MLLMs (Image models)} \\
\midrule
InternVL-1.5-26B & 42.43 & 5.75 & 14.82 & 14.19 & 37.14 & 29.66 & 17.83 & 11.59 & 21.68\\
InternVL-2.0-26B & 41.71 & 9.34 & 25.50 & 24.83 & 44.77 & 34.00 & 15.33 & 7.65 & 25.39\\
InternLM-XComposer-2.5 & 12.84 & 1.62 & 9.34 & 7.10 & 27.05 & 21.94 & 11.93 & 5.39 & 12.15\\
MiniCPM-V-2.6 & 49.17 & 10.05 & 18.34 & 23.06 & 46.57 & 35.44 & {\colorbox{green!20}{23.95}} & 11.99 & 27.32\\
Mantis-8B-Idefics2 & 18.25 & 0.18 & 23.04 & 20.73 & 0.92 & 1.06 & 0.24 & 0.11 & 8.07\\
Phi-3-vision-128k-instruct & 0.30 & 0.18 & 18.01 & 22.95 & 1.73 & 2.82 & 0.97 & 2.29 & 6.16\\
Qwen2-VL-7B & 14.93 & 1.97 & 24.83 & 24.72 & 9.55 & 13.83 & 4.29 & 4.26 & 12.30\\
LLaVA-NeXT-34B & 0.47 & 0.00 & 22.71 & 24.28 & 18.49 & 20.64 & 7.91 & 8.13 & 12.83\\
CogVLM-chat-hf & 3.50 & 0.18 & 22.71 & 25.94 & 1.75 & 1.90 & 0.29 & 0.22 & 7.06\\
Yi-VL-6B & 2.90 & 0.36 & 25.67 & 25.83 & 6.83 & 17.89 & 10.11 & 18.59 & 13.52\\
Qwen-VL-Chat & 2.47 & 1.26 & 15.16 & 18.85 & 1.75 & 13.91 & 5.13 & 8.89 & 8.43\\
Llama-3.2-11B-Vision & 9.85 & 0.18 & 22.04 & 23.28 & 4.69 & 9.20 & 3.07 & 6.28 & 9.82\\
\bottomrule
\end{tabu}
}
\caption{Experiment result on combined image (combine multiple images into one large image). Best results are marked in green.}
\label{result_single}%
\end{table*}

\subsection{Discussion}

\paragraph{Human evaluation.}
We select three models with good performance from the closed-source, open-source multi-image, and open-source video categories in Table~\ref{main_result} for human evaluation. Specifically, we randomly sample 100 examples from both the easy and hard versions of the Multi-image Extraction and Multi-image Reasoning tasks respectively, and invite four experts to score whether the predicted answers are faithful to the ground truth, with the scores divided into three levels: fully faithful (2 points), partially faithful (1 point), and not faithful at all (0 points). The average results are shown in Figure~\ref{table:Human_Evaluation_result}. From these results, we observe that Qwen-VL-Max achieves the best performance, followed by MiniCPM-V-2.6, with VideoLLaMA2-7B performing the worst, which is align with the results in Table~\ref{main_result}. The human evaluation and the metrics in Table~\ref{main_result} corroborate each other, further supporting our analysis conclusions.


\paragraph{Can models designed for single-image inputs also be used for multi-image understanding?}

To evaluate the performance of MLLMs that only support single-image input on our dataset, we combine multiple images from each data sample into one large image. This approach not only tests the performance of MLLMs that support multi-image input on the composite image but also assesses the capability of MLLMs that only support single-image input in understanding multiple images. Here, in addition to the previous support multi-image input models, we select 5 representative models that only support single-image input: LLaVA-NeXT-34B, CogVLM-chat-hf, Yi-VL-6B, Qwen-VL-Chat, and Llama-3.2-11B-Vision. The results on combined images are shown in Table~\ref{result_single}. Compared to the main results presented in Table~\ref{main_result}, we found that closed-source models supporting multi-image input generally perform worse on composite images than when directly inputting multiple individual images. One possible reason is that these closed-source models have limited training data on composite images. In contrast, open-source visual models designed for multi-image input tend to perform better with composite images compared to handling multiple separate images. This indicates that, although these open-source models are designed for multi-image input, they perform better when processing single images. Moreover, we found that open-source models supporting only single-image input generally perform worse on composite images than open-source models that support multi-image input. This is reasonable since open-source models that support multi-image input typically excel in single-image performance.

\paragraph{Additional Analysis.}



In \textbf{Appendix}~\ref{Additional_Analysis}, we provide the following additional analysis: (1) the performance of MLLMs on the unanswerable question set. and (2) we provide a comprehensive qualitative analysis as well as an error analysis.


\section{Conclusion}

In this paper, we present the first Chinese multi-image dataset, RealBench. This dataset is based on real user inputs, making it highly relevant to the real world. RealBench covers four distinct tasks: multi-image retrieval, ranking, extraction, and reasoning. We conduct a comprehensive evaluation of RealBench using 21 MLLMs. The experimental results indicate that even the most powerful closed-source models encounter limitations when handling multi-image scenarios in Chinese. Additionally, there is a substantial performance gap between open-source models and closed-source models. These findings highlight the importance of RealBench and encourage the community to explore multi-image understanding capabilities within the Chinese context. 


\section*{Limitations}
The current study has two limitations that warrant further attention. First, although we have evaluated 21 representative multimodal LLMs on the dataset, the current evaluation coverage is still not exhaustive. It remains necessary to incorporate a broader range of models in future evaluations. Second, while the dataset is already of considerable scale, it can be further expanded to meet the growing data demands of large-scale pretraining models and support more comprehensive evaluations.

\section*{Ethical Considerations}

In this study, we rigorously adhere to established ethical standards and provides robust protection for privacy. Throughout the data processing phase, we have implemented a comprehensive set of anonymization techniques designed to prevent the exposure of any personally identifiable information. This includes the careful removal or obfuscation of sensitive identifiers, such as names, addresses, phone numbers, ID numbers, and other potentially identifying features, that could be used to trace back to individuals. These procedures are in place to ensure that any data analyzed remains anonymous and cannot be linked to specific individuals without proper authorization. Furthermore, for facial images present in the dataset, we have gone a step further by applying advanced blurring techniques to ensure that any facial features are adequately obscured, minimizing the risk of inadvertent privacy violations. 

Besides, the collection and use of all data receive explicit consent from the users and data providers. Prior to the commencement of data collection, users are informed about the nature of the data being collected, the specific purposes for which it would be used, and the terms under which the data would be processed and stored. This informed consent process ensures that all participants are fully aware of their rights, and their participation is voluntary. We are unwavering in our commitment to strictly safeguarding user privacy and ensuring that all data is used in compliance with legal requirements and ethical guidelines for data protection.


\bibliography{custom}

\appendix

\newpage

\section{The Distribution of 36 Categories}\label{Distribution_Categories}

According to the statistics, the ``travel'' category has the highest proportion at 5.25\%, while the 'astrology' category has the lowest proportion at 1.36\%. The proportions of the other categories generally hover around 2.7\%, resulting in an overall balanced distribution. This balanced distribution across categories helps preserve thematic diversity.

\section{Chinese Version of RealBench}\label{ChineseRealBench}

\subsection{Chinese Example of Each Task}


As shown in Figure~\ref{fig:main_chinese}, we provide chinese version of each task in RealBench.

\subsection{Four Type of Multi-image Reasoning}

As shown in Figure~\ref{fig:reason}, we provide an example for each reasoning type in the multi-image reasoning.

\section{Prompt Format}\label{Prompt_Format}

\subsection{Prompt Format for Constructing Data}
As shown in Figure~\ref{fig:task1_construct_prompt},~\ref{fig:task3_construct_prompt} and ~\ref{fig:task4_construct_prompt}, we provide different prompt formats for different tasks to construct the data. 


\subsection{Prompt Format for Evaluating}

As shown in Figure~\ref{fig:task1_evaluate_prompt}, ~\ref{fig:task2_evaluate_prompt}, ~\ref{fig:task3_evaluate_prompt} and ~\ref{fig:task4_evaluate_prompt}, we provide the evaluation prompt formats corresponding to each task.

\section{Additional Analysis}\label{Additional_Analysis}

\paragraph{The performance of MLLMs on the unanswerable question set}
For multi-image extraction and multi-image reasoning tasks, we manually annotate 438 data samples, each of them contains a question and multiple relevant images. We categorize these 438 data samples as a collection of questions that cannot be answered. In other word, we cannot extract the answer to the question from the text within the images, nor can we infer the answer based on the content of the images. In such cases, the model's correct response is ``No relevant content mentioned''. To enable the model to respond accurately, we design a guiding prompt, with the specific format shown in Figure~\ref{fig:prompt_unanswer}. Ultimately, the model's responses can either be ``No relevant content mentioned'' or other generated answers, with only ``No relevant content mentioned'' being correct. By calculating the accuracy, we assess each model's self-recognition ability, that is, knowing what it does not know. The results are shown in Figure~\ref{fig:unanswerable}. We can see that among the closed-source models, GPT-4V has the highest accuracy. Although its performance surpasses that of most open-source models, it still falls short compared to InternVL-1.5, InternVL-2.0, and CogVLM2. This indicates that the latter models have two advantages over the closed-source ones: (1) they use a large Chinese multi-image dataset for training; and (2) they incorporate many unanswerable negative samples during data construction. In contrast, Claude 3.5 Sonnet, Mantis, Video-LLaMA-2, Valley, and Longvila show relatively poor performance, likely due to the low proportion of unanswerable Chinese negative samples in their training sets.

\begin{figure}[t]
\centering
\begin{tcolorbox}[title=Prompts for unanswerable questions set]
\scriptsize

Multimodal Multi-Image Extraction Task: You will be provided with a
set of images and a question. Please extract the answer to the question
from the text in the images.
\\
\\
Requirements:
\\
1. The answer must be extracted only from the text in the images.
Output the original text and return in the format: 'Answer: ..... '
\\
2. If no relevant information is found in the images, reply with 'No
relevant content mentioned. '
\\
\\
Here are the questions and images I provide:

\end{tcolorbox}
\caption{The prompts for unanswerable questions set.}
\label{fig:prompt_unanswer}
\end{figure}

\begin{figure*}[t]
  \centering
   \includegraphics[width=0.98\linewidth]{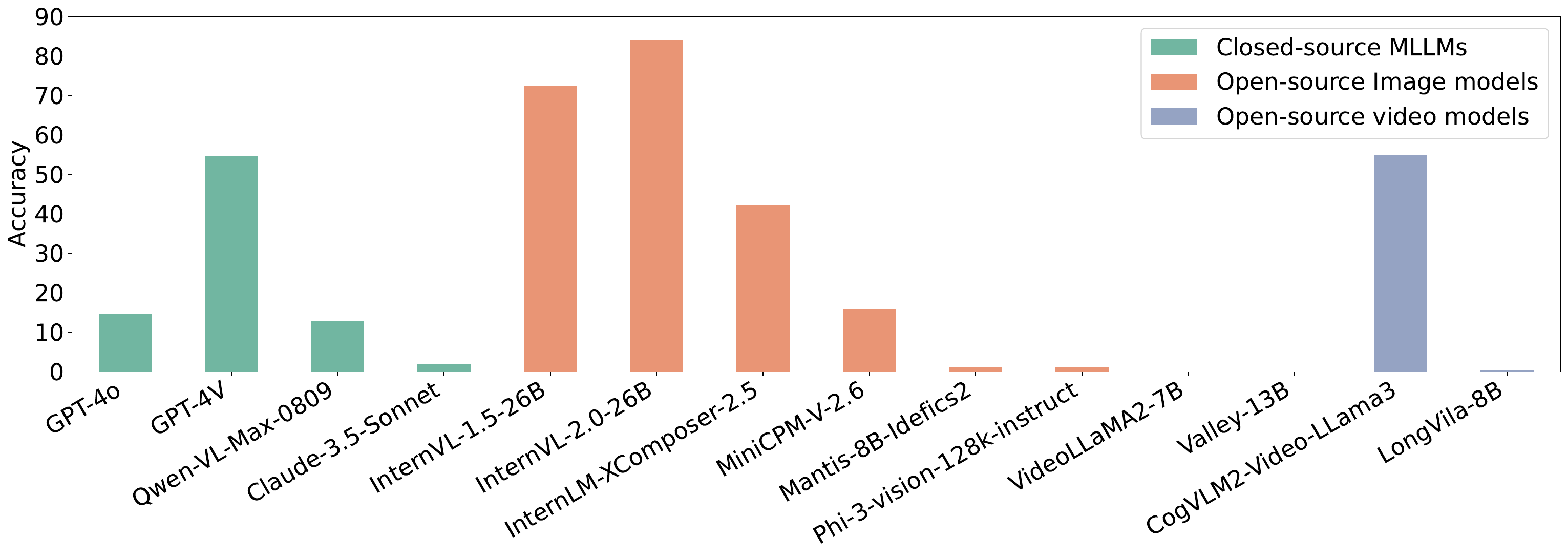}
   \caption{The performance of MLLMs on the unanswerable question set.}
   \label{fig:unanswerable}
\end{figure*}

\begin{table*}[t]
\centering
\setlength{\tabcolsep}{23pt}
\scalebox{0.85}{
\begin{tabu}{lcccc}
\toprule
Task Type & I & II & III & IV \\
\midrule
Ground Truth & 3 & C & Prince Henry Cliff Walk & egg yolk pastry \\
GPT-4V & 3 & C & Prince Henry Cliff Walk & egg yolk pastry \\
GPT-4o & 3 & C & Prince Henry Cliff Walk & egg yolk pastry \\
Claude-3.5-Sonnet & 3 & C & Prince Henry Cliff Walk & egg yolk pastry \\
InternVL-1.5-26B & 1 & - & - & durian crisp \\
InternVL-2.0-26B & 1 & A & - & durian crisp \\
Mantis-8B-Idefics2 & 5 & B & Image 8 & Image 9 \\
MiniCPM-V-2.6 & 3 & A & Prince Henry Cliff Walk & egg yolk pastry \\
Qwen2-VL-72B & 7 & A & Katoomba & egg tart \\
\bottomrule
\end{tabu}}
\caption{Answers from different models on the representative samples of the 4 tasks. The corresponding images and questions for the tasks are shown in Figure~\ref{fig:case}. The mark - indicates that the model fails to follow instructions to provide a valid answer.}
\label{tab:case}
\end{table*}

\begin{figure*}[t]
  \centering
   \includegraphics[width=0.95\linewidth]{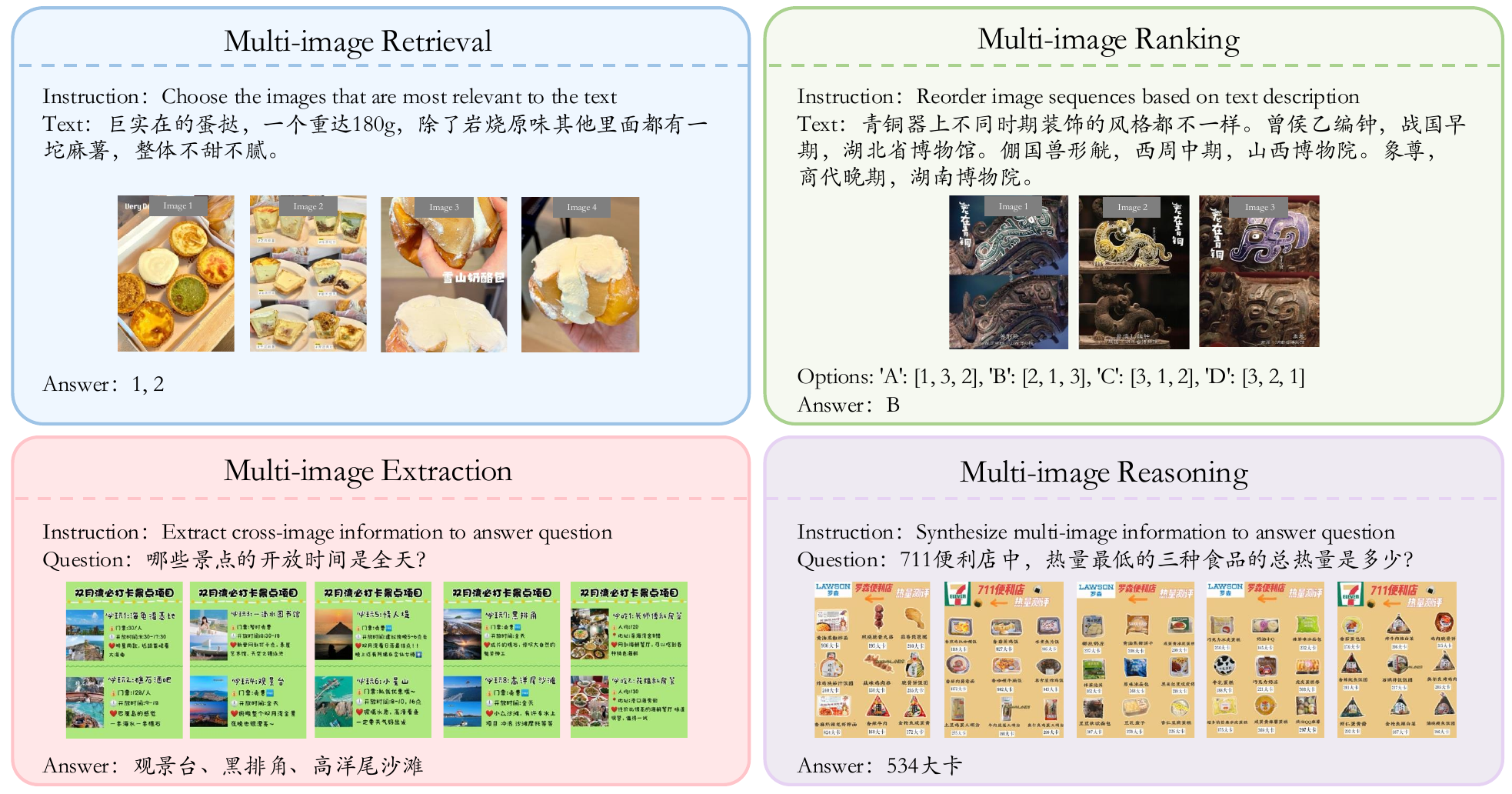}
   \caption{Detailed examples of the four tasks included in RealBench.}
   \label{fig:main_chinese}
\end{figure*}

\paragraph{Qualitative Analysis}
To provide a deeper understanding of the evaluation dataset and model performance, we conduct a qualitative analysis. For this purpose, we select a representative example from each task and compare the responses generated by each model, as presented in Table~\ref{tab:case} and Figure~\ref{fig:case}. Our findings indicate that all closed-source models answer the questions correctly, with MiniCPM performing the best among the open-source models. Notably, there is a substantial performance gap between closed-source and open-source models.


\paragraph{Error Analysis}
Through manual inspection, we identify four primary categories of errors (shown in Figure~\ref{fig:error}): (1) \textbf{Multimodal Alignment} In this type of error, the model fails to accurately link visual elements in the images to their corresponding textual descriptions. For example, when the text refers to ``peach'', the model should associate it with image 5, which contains peach segments. However, it incorrectly assigns image 3 as the answer; (2) \textbf{Text Recognition} The model fails to recognize and extract critical text embedded within the images. For instance, a target sentence appears clearly in image 2, yet the model incorrectly selects image 1 as the answer due to its inability to interpret the text accurately; (3) \textbf{Instruction Following} In some cases, the model fails to follow task instructions, producing responses that are unrelated to the context. For example, when instructed to answer a question, the model generates more questions instead of providing a valid answer, resulting in incorrect response; (4) \textbf{Knowledge} The model lacks essential knowledge, particularly commonsense, required to solve certain problems. For example, when asked to select a cake, the model incorrectly chooses an ice cream, likely due to a failure to understand that ice creams are not classified as cakes.





\begin{figure*}[t]
  \centering
   \includegraphics[width=0.95\linewidth]{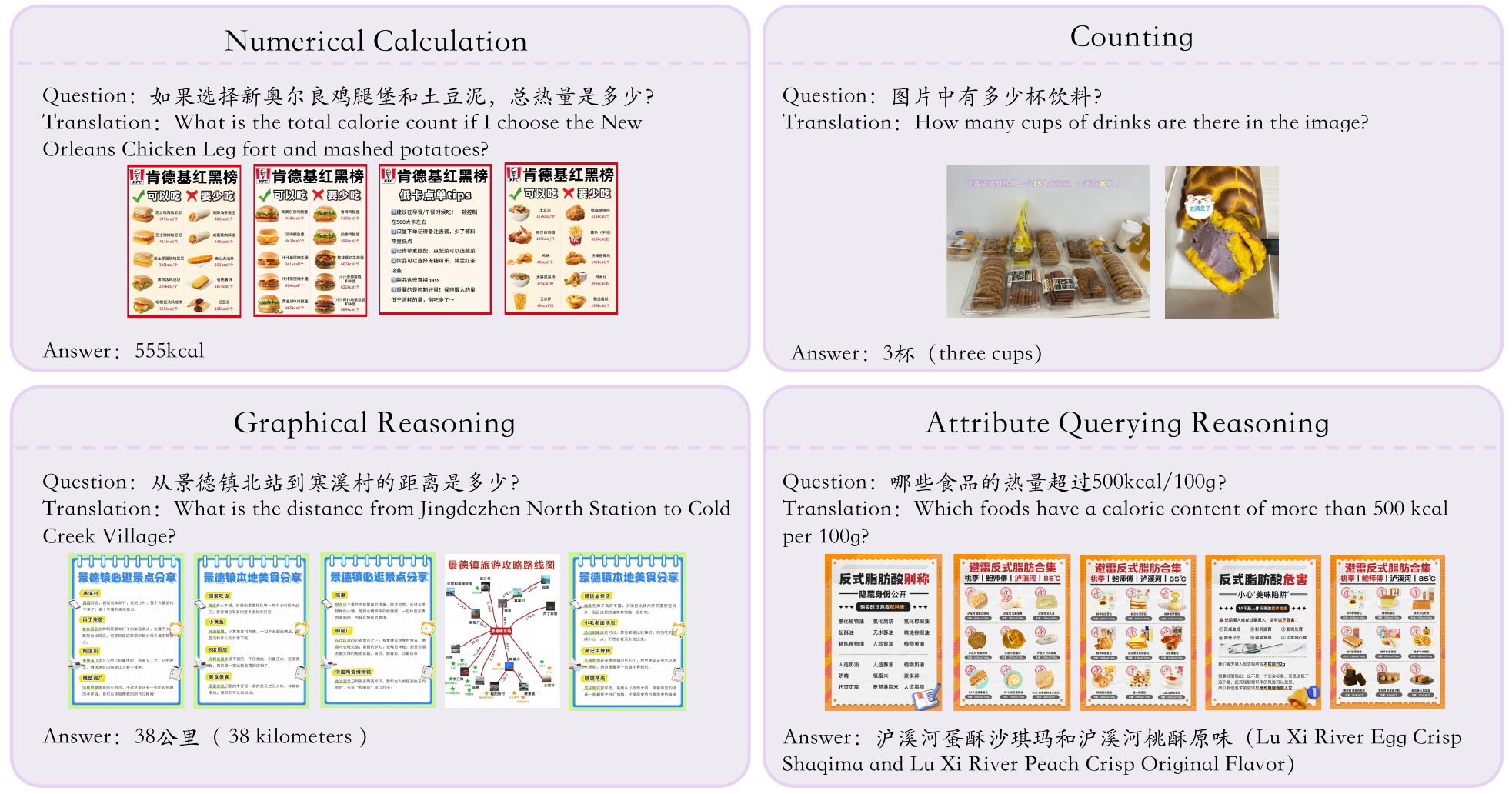}
   \caption{The examples of numerical calculation, counting, graphical reasoning, and attribute querying reasoning in the multi-image reasoning task.}
   \label{fig:reason}
\end{figure*}


\begin{figure*}[t]
\centering
\begin{tcolorbox}[title=Prompts for constructing the multi-image retrieval task]
\scriptsize

PROMPT\_TEMPLATE = """
\\
\\
\#\# Profile:
\\
- Role: Information Extraction Expert
\\
- Language: Chinese
\\
- Description: Effectively identify and extract the relationship between
text content and image numbers in notes composed of long text and
multiple images, and output the information in JSON format.
\\
\\
\#\# Goals:
\\
- Identify and extract image-related descriptive information from long
paragraphs of text
\\
- Extract the image numbers associated with the descriptive information
\\
- Output the identified and extracted information in JSON format
\\
\\
\#\# OutputFormat:
\\

[\{"text": "<extracted text content>", "image\_ids": [<corresponding image number list>]\}, \{"text": "<extracted text content>", "image\_ids": [<corresponding image number list>]\}]
\\
\\
\#\# Constraints:
\\
- Strictly follow the output format
\\
- No explanations are provided
\\
- The "text" section should not include "Image x", but only the information
related to the image
\\
- Text without a corresponding image should not be output
\\
\\
\#\# Input note text: 
\\
{text}"""

\end{tcolorbox}
\caption{The prompts used for constructing the multi-image retrieval task.}
\label{fig:task1_construct_prompt}
\end{figure*}

\begin{figure*}
  \centering
  \begin{subfigure}{0.48\linewidth}
    \begin{tcolorbox}[title=Prompt for easy version]
\scriptsize

Task Description: You will be responsible for creating questions and
answers based on a given set of images. Each image will correspond to
one question and one answer. You need to ensure a high level of
relevance between the question and the answer, while meeting the
following requirements:
\\
\\
1.Relevance between the question and the image: Each question must
be directly related to the content of the image and cannot be
detached from it.
\\
2.Clarity and conciseness of the question: The question should be
simple, clear, and directly point to key information in the image.
\\
3.Source of the answer: The answer must be entirely based on the text
information within the image, without including any content
outside of the image's text.
\\
4.Format: Each question and answer pair should be presented in the
following format: Question: ... Answer: ...

\end{tcolorbox}
    \caption{Prompt for easy version.}
    \label{fig:short-a}
  \end{subfigure}
  \hfill
  \begin{subfigure}{0.48\linewidth}
    \begin{tcolorbox}[title=Prompt for hard version]
\scriptsize

Task Description: You will be responsible for creating questions and
answers based on a given set of images. You need to ensure a high level
of relevance between the question and the answer, while meeting the
following requirements:
\\
\\
1.Relevance between the question and the images: Each question must
be directly related to the content of multiple images, ensuring a
close connection between the question and the images' content.
\\
2.Complexity of the question: The question should be designed to be
relatively complex, requiring the integration of information from
multiple images to construct.
\\
3.Source of the answer: The answer must be derived from the text
information within the multiple images, ensuring both accuracy and
comprehensiveness.
\\
4.Format: Each question and answer pair should be presented in the
following format: Question: ... Answer: ...

\end{tcolorbox}
    \caption{Prompt for hard version.}
    \label{fig:short-b}
  \end{subfigure}
  \caption{The prompts used for constructing the multi-image extraction task.}
  \label{fig:task3_construct_prompt}
\end{figure*}

\begin{figure*}
  \centering
  \begin{subfigure}{0.48\linewidth}
    \begin{tcolorbox}[title=Prompt for easy version]
\scriptsize

Task Description: You will be responsible for performing an image-based
reasoning task. Based on the provided set of images, you will construct a
series of questions and answers that require reasoning. The following
requirements must be met:
\\
\\
1.Relevance between the question and the images: Each question must
be directly related to the content of the images and cannot be
detached from the images.
\\
2.Clarity and conciseness of the question: The question should be
simple, clear, and directly point to key information within the
image.
\\
3.Diversity of reasoning: The questions should cover different types of
reasoning, including numerical calculations, counting, visual
reasoning, or attribute-based queries.
\\
4.Format: Each question and answer pair should be presented in the
following format: Question: ... Answer: ...

\end{tcolorbox}
    \caption{Prompt for easy version.}
    \label{fig:short-a}
  \end{subfigure}
  \hfill
  \begin{subfigure}{0.48\linewidth}
    \begin{tcolorbox}[title=Prompt for hard version]
\scriptsize

Task Description: You will be responsible for creating questions and
answers based on a given set of images. You need to ensure a high level
of relevance between the question and the answer, while meeting the
following requirements:
\\
\\
1.Relevance between the question and the images: Each question must
be directly related to the content of multiple images, ensuring a
close connection between the question and the images.
\\
2.Complexity of the question: The question should be designed to be
relatively complex, requiring the integration of information from
multiple images to construct.
\\
3.Source of the answer: The answer must be derived from the text
information within the multiple images, ensuring accuracy and
comprehensiveness.
\\
4.Format: Each question and answer pair should be presented in the
following format: Question: ... Answer: ...

\end{tcolorbox}
    \caption{Prompt for hard version.}
    \label{fig:short-b}
  \end{subfigure}
  \caption{The prompts used for constructing the multi-image reasoning task.}
  \label{fig:task4_construct_prompt}
\end{figure*}

\begin{figure*}
  \centering
  \begin{subfigure}{0.48\linewidth}
    \begin{tcolorbox}[title=Prompt for easy version]
\scriptsize

You will be given a piece of text and a set of images. Your task is to
identify and list the image number most relevant to the given text. There is
only one relevant image, and the numbering starts from 1. Only return the
image number, with no additional information.

\end{tcolorbox}
    \caption{Prompt for easy version.}
    \label{fig:short-a}
  \end{subfigure}
  \hfill
  \begin{subfigure}{0.48\linewidth}
    \begin{tcolorbox}[title=Prompt for hard version]
\scriptsize

You will be given a piece of text and a set of images. Your task is to
identify and list the image numbers most relevant to the given text. There
may be multiple relevant images, and the numbering starts from 1. Only
return the image numbers, separated by commas, with no additional
information.

\end{tcolorbox}
    \caption{Prompt for hard version.}
    \label{fig:short-b}
  \end{subfigure}
  \caption{The prompts used for evaluating the multi-image retrieval task.}
  \label{fig:task1_evaluate_prompt}
\end{figure*}

\begin{figure*}[t]
  \centering
   \begin{tcolorbox}[title=Prompt for evaluating the multi-image ranking task]
\scriptsize

You will play the role of an image-text note analysis expert. You will
receive an image-text note, which includes a paragraph of text with
missing image numbers and a set of unordered images, as well as
multiple image order options. The text part uses 'Figure\_' to represent
missing image numbers. Based on the image-text information, you need
to sequentially infer the image number for each ' Figure\_' in the given
image sequence (starting from 1). Then, you should select the correct
image order from the options and return the corresponding letter. Your
answer should consist of only one letter, with no additional information.

\end{tcolorbox}
   \caption{The prompts used for evaluating the multi-image ranking task.}
   \label{fig:task2_evaluate_prompt}
\end{figure*}

\begin{figure*}
  \centering
  \begin{subfigure}{0.48\linewidth}
    \begin{tcolorbox}[title=Prompt for easy version]
\scriptsize

You will be given a set of images and a question. Please extract the
answer to the question from one of the images, based on the text
information in that image.
\\
\\
Requirements:
\\
1.The answer must be accurate and can only be extracted from the
content of the image. >Output the original content. Return the format:
'The answer is: ...... '
\\
2.If the image contains multiple points of answer, display them
separated by line breaks.
\\
\\
Below are the question and images I provide:

\end{tcolorbox}
    \caption{Prompt for easy version.}
    \label{fig:short-a}
  \end{subfigure}
  \hfill
  \begin{subfigure}{0.48\linewidth}
    \begin{tcolorbox}[title=Prompt for hard version]
\scriptsize

You will be given a set of images and a question. Please extract the
answer to the question from the set of images, based on the text
information from multiple images.
\\
\\
Requirements:
\\
1.The answer must be accurate and can only be extracted from the
content of the image. >Output the original content. Return the format:
'The answer is: ...... '
\\
2.If the image contains multiple points of answer, display them
separated by line breaks.
\\
\\
Below are the question and images I provide:

\end{tcolorbox}
    \caption{Prompt for hard version.}
    \label{fig:short-b}
  \end{subfigure}
  \caption{The prompts used for evaluating the multi-image extraction task.}
  \label{fig:task3_evaluate_prompt}
\end{figure*}

\begin{figure*}
  \centering
  \begin{subfigure}{0.48\linewidth}
    \begin{tcolorbox}[title=Prompt for easy version]
\scriptsize

Multimodal Reasoning Task: You will be given a set of images and a
question. Please derive the answer to the question from one of the
images. The answer should be inferred based on the text information
within that image. The reasoning types include numerical calculations,
counting, visual reasoning, or attribute-based queries.
\\
\\
Requirements:
\\
1.Output the correct answer. Return the format: 'The answer is: .....'
\\
2.If the image contains multiple points of answer, display them
separated by line breaks.
\\
\\
Below are the question and images I provide:

\end{tcolorbox}
    \caption{Prompt for easy version.}
    \label{fig:short-a}
  \end{subfigure}
  \hfill
  \begin{subfigure}{0.48\linewidth}
    \begin{tcolorbox}[title=Prompt for hard version]
\scriptsize

Multimodal Reasoning Task: You will be given a set of images and a
question. Please derive the answer to the question from the set of images.
The answer should be inferred based on the text information across
multiple images. The reasoning types include numerical calculations,
counting, visual reasoning, or attribute-based queries.
\\
\\
Requirements:
\\
1.Output the correct answer. Return the format: 'The answer is: .....'
\\
2.If the images contain multiple points of answer, display them
separated by line breaks.
\\
\\
Below are the question and images I provide:

\end{tcolorbox}
    \caption{Prompt for hard version.}
    \label{fig:short-b}
  \end{subfigure}
  \caption{The prompts used for evaluating the multi-image reasoning task.}
  \label{fig:task4_evaluate_prompt}
\end{figure*}

\begin{figure*}[t]
  \centering
   \includegraphics[width=\linewidth]{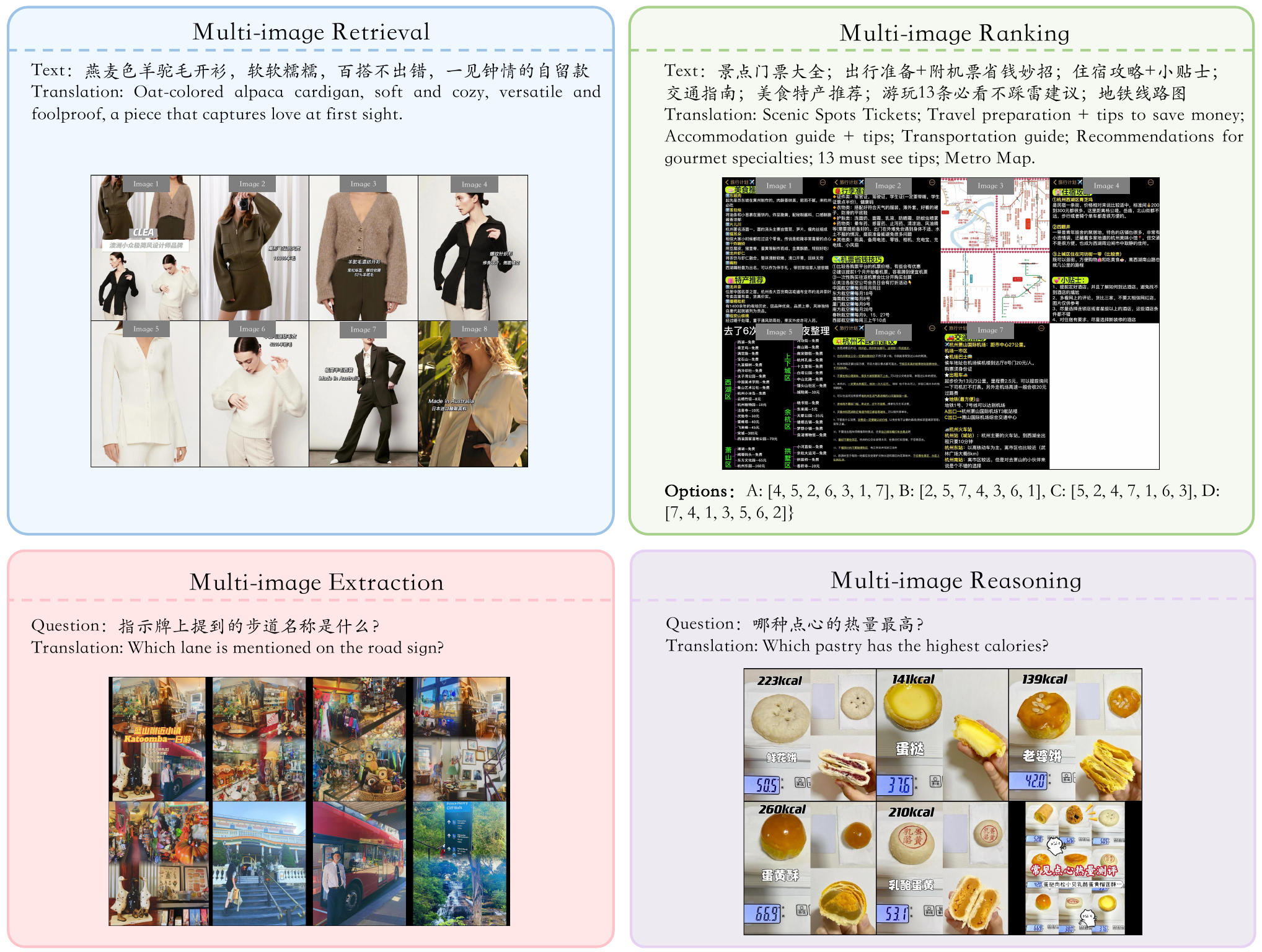}
   \caption{Images and context for the 4 representative samples from each task.}
   \label{fig:case}
\end{figure*}

\begin{figure*}[t]
  \centering
   \includegraphics[width=\linewidth]{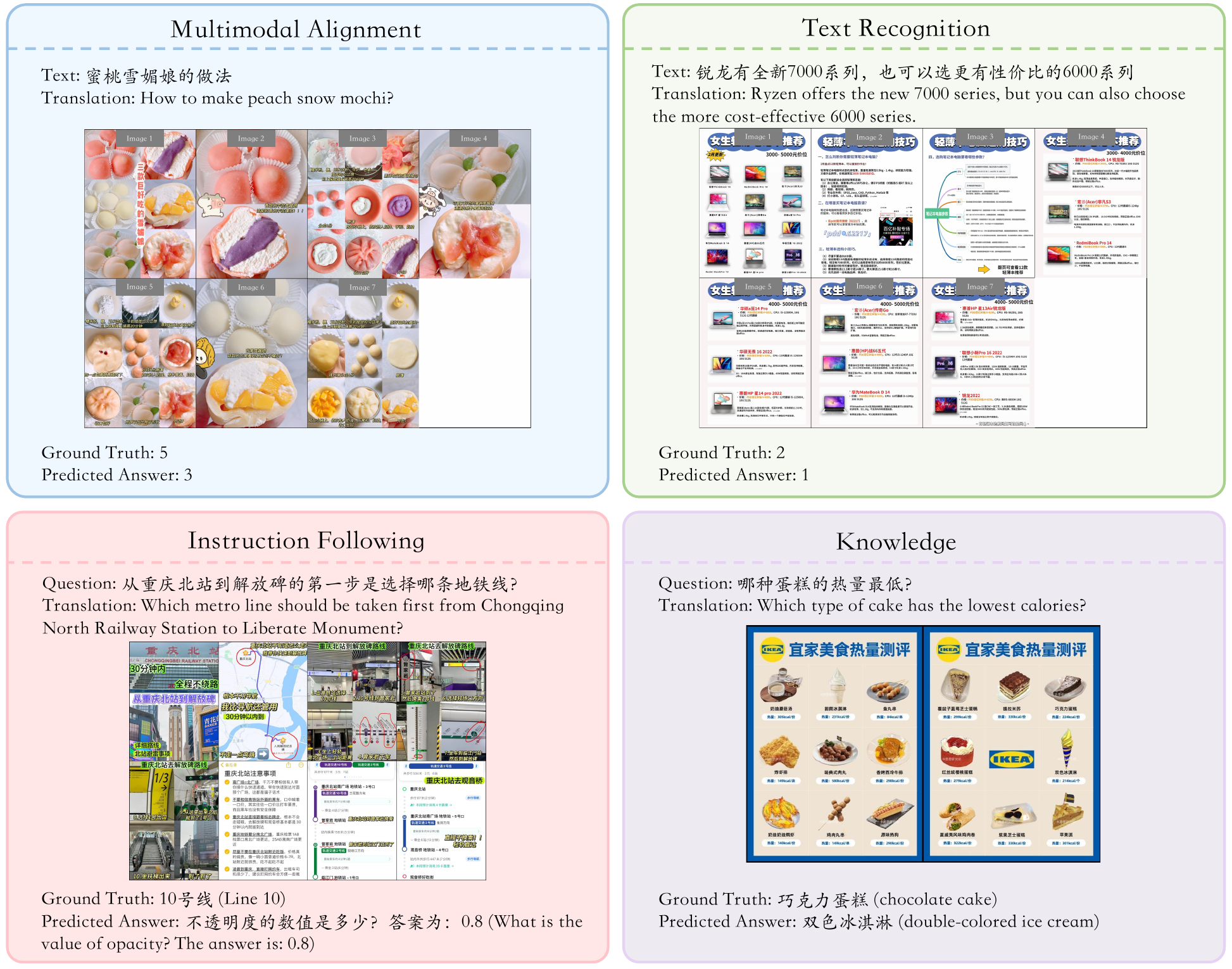}
   \caption{Common types of errors identified.}
   \label{fig:error}
\end{figure*}

\end{document}